\documentclass{article}

\usepackage[preprint]{neurips_2022}

\usepackage{xcolor}         
\definecolor{linkColor}{rgb}{0.18,0.39,0.62}
\usepackage[utf8]{inputenc} 
\usepackage[T1]{fontenc}    
\usepackage[colorlinks=true,linkcolor=linkColor,citecolor=linkColor,filecolor=linkColor,urlcolor=linkColor]{hyperref}       
\usepackage{url}            
\usepackage{booktabs}       
\usepackage{amsfonts}       
\usepackage{nicefrac}       
\usepackage{microtype}      
\usepackage[most]{tcolorbox}
\usepackage{enumitem}

\usepackage{amsmath}
\usepackage{amssymb}
\usepackage{mathtools}
\usepackage{amsthm}
\usepackage{graphicx}
\usepackage{subfigure}
\usepackage{xspace}
\usepackage{pifont}
\usepackage{xcolor}
\usepackage{array}

\theoremstyle{plain}

\theoremstyle{definition}

\theoremstyle{remark}

\usepackage[capitalize,noabbrev]{cleveref}
\usepackage{framed}
\usepackage{multirow}
\usepackage{svg}
\usepackage{listings}
\usepackage{wrapfig}

\usepackage{multirow}
\usepackage{fontawesome5}

\newcommand\our{\textsc{VALL-E}}

\newcommand{\cmark}{\ding{51}\xspace}%
\newcommand{\xmark}{\ding{55}\xspace}%

\makeatletter

\newcommand*{\@rowstyle}{}

\newcommand*{\rowstyle}[1]{
  \gdef\@rowstyle{#1}%
  \@rowstyle\ignorespaces%
}

\newcolumntype{=}{
  >{\gdef\@rowstyle{}}%
}

\newcolumntype{+}{
  >{\@rowstyle}%
}

\makeatother

\title{Neural Codec Language Models are \\ Zero-Shot Text to Speech Synthesizers}

\author{%
Chengyi Wang$^*$   \ Sanyuan Chen$^*$  \ Yu Wu$^*$  Ziqiang Zhang \ Long Zhou \ Shujie Liu  \\ \textbf{Zhuo Chen \ Yanqing Liu \ Huaming Wang \ Jinyu Li   \ Lei He \ Sheng Zhao \ Furu Wei }
\\ Microsoft
\\ \url{https://github.com/microsoft/unilm}
}

\begin{document}

\maketitle
\def\thefootnote{$*$}\footnotetext{These authors contributed equally to this work. Correspondence: \{yuwu1,shujliu,fuwei\}@microsoft.com}

\begin{abstract}

We introduce a language modeling approach for text to speech synthesis (TTS). 
Specifically, we train a \textit{neural codec language model} (called \our{}) using discrete codes derived from an off-the-shelf neural audio codec model, and regard TTS as a conditional language modeling task rather than continuous signal regression as in previous work.
During the pre-training stage, we scale up the TTS training data to 60K hours of English speech which is hundreds of times larger than existing systems.
\our{} emerges \textit{in-context learning} capabilities and can be used to synthesize high-quality personalized speech with only a 3-second enrolled recording of an unseen speaker as an acoustic prompt. Experiment results show 
 that \our{} significantly outperforms the state-of-the-art zero-shot TTS system  in terms of speech naturalness and speaker similarity. In addition, we find \our{} could preserve the speaker's emotion and acoustic environment of the acoustic prompt in synthesis.  See \url{https://aka.ms/valle} for demos of our work. 





\end{abstract}

\begin{figure*}[h]
	\centering
	\includegraphics[width=0.88\textwidth]{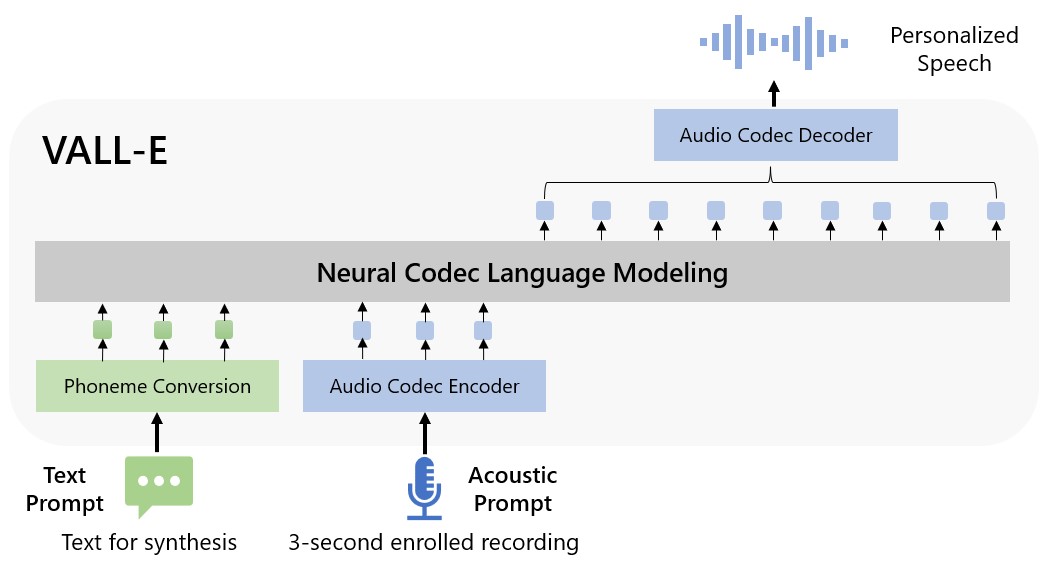}
	\caption{ The overview of \our{}. Unlike the previous pipeline (e.g., phoneme $\rightarrow$ mel-spectrogram $\rightarrow$ waveform), the pipeline of \our{} is phoneme $\rightarrow$ discrete code $\rightarrow$ waveform.  \our{} generates the discrete audio codec codes based on phoneme and acoustic code prompts, corresponding to the target content  and the  speaker's voice. \our{} directly enables various speech synthesis applications, such as zero-shot TTS, speech editing, and content creation combined with other generative AI models like GPT-3 \citep{DBLP:conf/nips/BrownMRSKDNSSAA20}.
    }\label{overview}
\end{figure*}%

\newpage
\section{Introduction}

 The last decade has yielded dramatic breakthroughs in speech synthesis through the development of neural networks and end-to-end modeling. Currently, cascaded text to speech (TTS) systems \citep{DBLP:conf/icassp/ShenPWSJYCZWRSA18, DBLP:conf/nips/RenRTQZZL19,DBLP:conf/aaai/Li0LZL19} usually leverage a pipeline with an acoustic model and a vocoder using mel spectrograms as the intermediate representations. 
While advanced TTS systems can synthesize high-quality speech from single or multiple speakers \citep{DBLP:conf/interspeech/LiuXH0Z22, DBLP:conf/icml/KimKS21}, it still requires high-quality clean data from the recording studio. Large-scale data crawled from the Internet cannot meet the requirement, and always lead to performance degradation. Because the training data is relatively small, current TTS systems still suffer from poor generalization. Speaker similarity and speech naturalness decline dramatically for unseen speakers in the zero-shot scenario. 
 To tackle the zero-shot TTS problem, existing work leverages speaker adaptation \citep{DBLP:conf/iclr/ChenASBRZWCTLGO19, DBLP:conf/interspeech/WangTFYWZ20} and speaker encoding  \citep{DBLP:conf/nips/ArikCPPZ18,casanova2022yourtts} methods, requiring additional fine-tuning, complex pre-designed features, or heavy structure engineering. 

Instead of designing a complex and specific network for this problem, the ultimate solution is to train a model with large and diverse data as much as possible, motivated by success in the field of text synthesis \citep{DBLP:conf/nips/BrownMRSKDNSSAA20,DBLP:journals/corr/abs-2204-02311}.  
Recent years have witnessed  notable performance improvement for data increase in the text language model, from 16GB of uncompressed text \citep{devlin2019bert}, to 160GB \citep{liu2019roberta}, to 570GB \citep{DBLP:conf/nips/BrownMRSKDNSSAA20}, and finally, around 1TB \citep{DBLP:journals/corr/abs-2204-02311}. 
Transferring this success to the field of speech synthesis, 
we introduce \our{}, the first language model based TTS framework leveraging the large, diverse, and multi-speaker speech data. 
As shown in Figure \ref{overview}, to synthesize personalized speech (e.g., zero-shot TTS), \our{} generates the corresponding acoustic tokens conditioned on the acoustic tokens of the 3-second enrolled recording and the phoneme prompt, which  constrain the speaker and content information respectively.  
Finally, the generated acoustic tokens are used to synthesize the final waveform with the corresponding neural codec decoder  \citep{defossez2022high}.  
The discrete acoustic tokens  derived from an audio codec model  enable us to  treat TTS as conditional codec language modeling, and advanced prompting-based large-model techniques (as in GPTs \citep{DBLP:conf/nips/BrownMRSKDNSSAA20}) can be leveraged for the TTS tasks. The acoustic tokens also allow us to generate diverse synthesized results in TTS by using different sampling strategies during inference.

 We train \our{} with LibriLight \citep{kahn2020libri}, a corpus consisting of 60K hours of English speech with over 7000 unique speakers. The original data is audio-only, so we employ a speech recognition model to generate the transcriptions.  Compared to previous TTS training datasets, such as LibriTTS \citep{DBLP:conf/interspeech/ZenDCZWJCW19},  our data contain more noisy speech and inaccurate transcriptions but provide diverse speakers and prosodies. We believe the proposed approach is robust to the noise and generalize well by leveraging large data. It is worth noting that existing TTS systems are always trained with dozens of hours of single-speaker data or hundreds of hours of multi-speaker data, which is over hundreds of times smaller than \our{}. 
 Table \ref{adv} summarizes the innovation of \our{}, a  language model approach for TTS, using audio codec codes as intermediate representations,  leveraging large and diverse data, leading to strong in-context learning capabilities. 


\begin{table}[h]
\begin{center}
\caption{A comparison between \our{} and current cascaded TTS systems.}
\begin{tabular}{c|c|c}
\toprule
& \bf Current Systems  & \textbf{\our{}} \\  \hline
Intermediate representation & mel spectrogram   &   audio codec code \\ \hline
Objective function &  continuous signal regression & language model \\ \hline
Training data & $\leq 600$ hours   & 60K hours   \\ \hline
In-context learning & \xmark   & \cmark  \\
\bottomrule
\end{tabular}
\label{adv}
\end{center}
\end{table}


We evaluate \our{} on LibriSpeech \citep{panayotov2015librispeech} and VCTK \citep{veaux2016superseded} datasets, where all test speakers are unseen in the training corpus. \our{} significantly outperforms the state-of-the-art zero-shot TTS system 
 \citep{casanova2022yourtts} in terms of speech naturalness and speaker similarity, with +0.12 comparative mean option score (CMOS) and +0.93 similarity mean option score (SMOS) improvement on LibriSpeech. 
\our{} also beats the baseline on VCTK with +0.11 SMOS and +0.23 CMOS improvements. It even achieves a +0.04 CMOS score against ground truth, showing the synthesized speech of unseen speakers is as natural as human recordings on VCTK. Moreover, the qualitative analysis shows that \our{} is able to synthesize diverse outputs with the same text and target speaker, which could benefit pseudo-data creation for the speech recognition task. We also find that \our{} could keep the acoustic environment (e.g.,  reverberation) and emotion (e.g. anger) of the acoustic prompt. 

In summary, we make the following contributions.
\begin{itemize}
  \item We propose \our{}, the first TTS framework with strong in-context learning capabilities as GPT-3, which treats TTS as a language model task with audio codec codes as an intermediate representation to replace the traditional mel spectrogram. It has in-context learning capability and enables prompt-based approaches for zero-shot TTS, which does not require additional structure engineering, pre-designed acoustic features, and fine-tuning as in previous work.  
  \item We  build a generalized TTS system in the speaker dimension by leveraging a huge amount of semi-supervised data, suggesting that simple scaling up semi-supervised data has been underestimated for TTS. 
  \item \our{} is able to provide  diverse outputs with the same input text and keep the acoustic environment and speaker's emotion of the acoustic prompt. 
  \item  We verify that \our{} synthesizes natural speech with high speaker similarity by prompting in the zero-shot scenario. Evaluation results show that \our{}  significantly outperforms the state-of-the-art zero-shot TTS system on LibriSpeech and VCTK. 
\end{itemize}
We encourage the reader to listen to our samples on the demo page \url{https://aka.ms/valle}.
\section{Related Work}
\textbf{Zero-Shot TTS:} Current TTS methods can be categorized into cascaded  and end-to-end methods. Cascaded TTS systems \citep{DBLP:conf/icassp/ShenPWSJYCZWRSA18, DBLP:conf/nips/RenRTQZZL19,DBLP:conf/aaai/Li0LZL19} usually leverage a pipeline with an acoustic model and a vocoder using mel spectrograms as the intermediate representations. To tackle the drawbacks of the vocoder, end-to-end TTS models \citep{DBLP:conf/icml/KimKS21, DBLP:conf/interspeech/LiuXH0Z22} are proposed to jointly optimize the acoustic model and vocoder. 
  In real scenarios, it is highly desirable to customize a TTS system to an arbitrary voice with rare enrolled recordings.
Therefore, there is growing interest in the zero-shot multi-speaker TTS techniques, and most of work is done in the context of cascaded TTS systems. 
 As the pioneers, \cite{DBLP:conf/nips/ArikCPPZ18} proposes speaker adaptation and speaker encoding approaches. In the line of speaker adaptation, the following work \citep{DBLP:conf/iclr/ChenASBRZWCTLGO19, DBLP:conf/interspeech/WangTFYWZ20,DBLP:conf/iclr/Chen0LLQZL21} tries to improve the adaptation efficiency with less target speaker data and speaker-specific parameters. \cite{DBLP:journals/taslp/HuangLLCL22} applies meta-learning on speaker adaptation, which only requires 5-shot to build a well-performed system. In parallel, speaker encoding-based methods achieved great progress in recent years.  A speaker encoding based system contains a speaker encoder and a TTS component, where the speaker encoder could be pre-trained on the speaker verification task \citep{DBLP:conf/nips/JiaZWWSRCNPLW18}. In \cite{DBLP:conf/nips/JiaZWWSRCNPLW18} and \cite{DBLP:conf/nips/ArikCPPZ18}, the experiments show that the model is able to generate high-quality outputs with 3 seconds enrolled recordings for in-domain speakers. To improve the quality of unseen speakers, advanced speaker embedding models \citep{DBLP:conf/odyssey/CaiCL18} can be employed, but it is still undesirable according to \cite{DBLP:journals/corr/abs-2106-15561}. Another way is to design advanced but complex speaker encoder \citep{DBLP:conf/interspeech/Wu00HZSQL22}. Diffusion model based TTS \citep{DBLP:conf/icml/PopovVGSK21, DBLP:conf/icml/KimKY22} is also extended to zero-shot TTS  \citep{DBLP:journals/corr/abs-2211-09383} and achieved good results.  Compared to previous work \citep{DBLP:conf/nips/RenRTQZZL19, DBLP:conf/interspeech/DuGC022}, our work follows the line of cascaded TTS but first uses audio codec code as intermediate representations. It is the first one that has strong in-context learning capabilities as GPT-3, which does not require fine-tuning, pre-designed features, or a complex speaker encoder.

\textbf{Spoken generative pre-trained models:} Self-supervised learning is widely investigated in the field of speech understanding \citep{baevski2020wav2vec,hsu2021hubert,chen2022wavlm} and speech-to-speech generation \citep{DBLP:journals/corr/abs-2102-01192, DBLP:journals/corr/abs-2209-03143}. In the context of speech-to-speech generation, a hot topic is how to synthesize speech in a textless setting. GSLM \citep{DBLP:journals/corr/abs-2102-01192} proposes to synthesize speech based on  HuBERT codes \citep{hsu2021hubert}, and \cite{DBLP:conf/interspeech/PolyakACKLHMD21} improves the performance by combining HuBERT codes with codes of VQVAE and a speaker encoder. 
AudioLM \citep{DBLP:journals/corr/abs-2209-03143} follows a similar way but use audio codecs \citep{DBLP:journals/taslp/ZeghidourLOST22} to synthesize speech, together with semantic codes. It should be noted that AudioLM is able to synthesize speech based on audio codecs without training an additional vocoder such as HifiGAN \citep{DBLP:conf/nips/KongKB20}. AudioLM is a speech-to-speech model, whereas \our{} is a TTS model, so we can explicitly control the content in speech synthesis.  
Another direction is to apply pre-training to the neural TTS. \cite{DBLP:conf/icassp/ChungWHZS19} pre-trains speech decoder in TTS through autoregressive mel-spectrogram prediction. 
In \cite{ao2022speecht5}, the authors propose a unified-modal encoder-decoder framework SpeechT5, which can leverage unlabeled speech and text data to pre-train all components of TTS model.
\cite{DBLP:conf/interspeech/TjandraS0S0019} quantizes unlabeled speech into discrete tokens by a VQVAE model \citep{van2017neural}, and train a model with the token-to-speech sequence. They demonstrate that the pre-trained model only requires a small amount of real data for fine-tuning. \cite{DBLP:conf/icml/BaiZCML022} proposes mask and reconstruction on mel spectrogram and showing better performance on speech editing and synthesis. Previous TTS pre-training work leverages less than 1K hours of data, whereas \our{} is pre-trained with 60K hours of data. Furthermore, \our{} is the first to use audio codec codes as  intermediate representations, and emerge in-context learning capability in zero-shot TTS.


\begin{figure*}[t]
\includegraphics[width=1.0\linewidth]{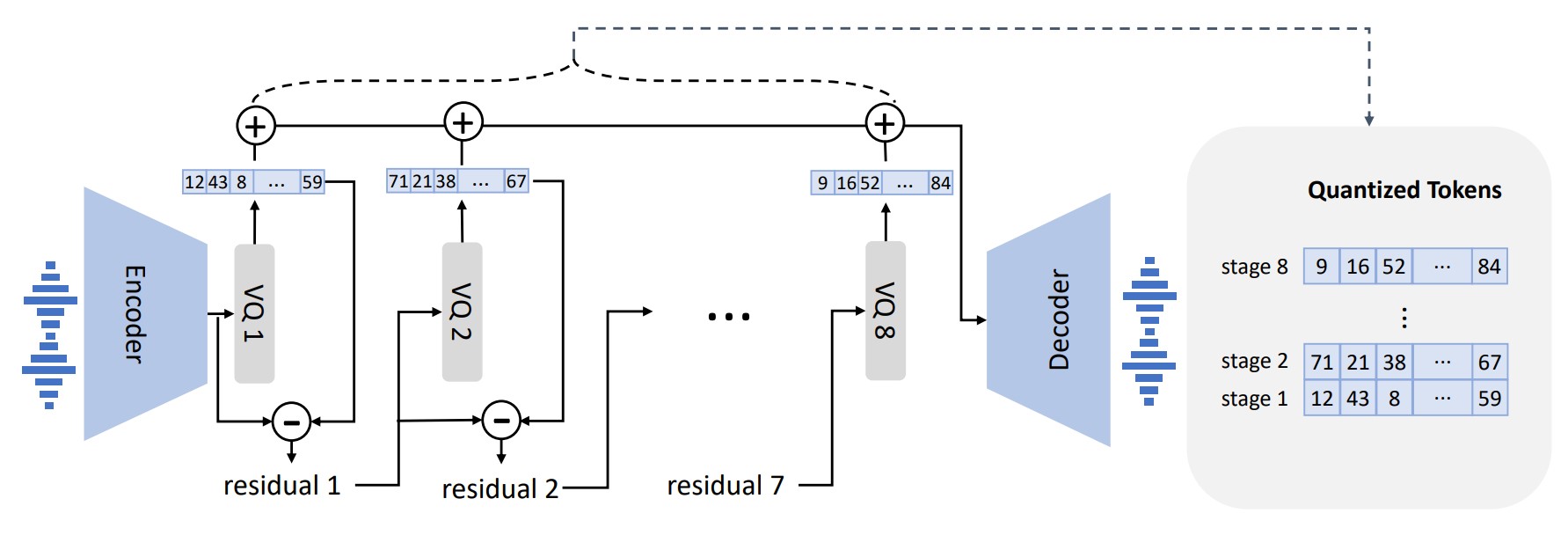} 
\caption{The neural audio codec model revisit. Because RVQ is employed, the first quantizer plays the most important role in reconstruction, and the impact from others gradually decreases.\label{codec} }

\end{figure*}
\section{Background: Speech Quantization}
Since audio is typically stored as a sequence of 16-bit integer values, a generative model is required to output $2^{16}=65,536$ probabilities per timestep to synthesize the raw audio. In addition, the audio sample rate exceeding ten thousand leads to an extraordinarily long sequence length, making it more intractable for raw audio synthesis. 
To this end, speech quantization is required to compress integer values and sequence length.
$\mu$-law transformation can quantize each timestep to 256 values and reconstruct high-quality raw audio. It is widely used in speech generative models, such as WaveNet \citep{DBLP:conf/ssw/OordDZSVGKSK16}, but the inference speed is still slow since the sequence length is not reduced. Recently, vector quantization is widely applied in self-supervised speech models for feature extraction, such as vq-wav2vec \citep{DBLP:conf/iclr/BaevskiSA20} and HuBERT \citep{hsu2021hubert}. The following work \citep{DBLP:journals/corr/abs-2102-01192,DBLP:conf/interspeech/DuGC022} shows the codes from self-supervised models can also reconstruct content, and the inference speed is faster than WaveNet. However, the speaker identity has been discarded and the reconstruction quality is low \citep{DBLP:journals/corr/abs-2209-03143}.  AudioLM \citep{DBLP:journals/corr/abs-2209-03143} trains speech-to-speech language models on both k-means tokens from a self-supervised model and acoustic tokens from a neural codec model, leading to high-quality speech-to-speech generation. 

 In this paper, we follow AudioLM \citep{DBLP:journals/corr/abs-2209-03143} to leverage neural codec models to represent speech in discrete tokens. To compress audio for network transmission, codec models are able to encode waveform into discrete  acoustic codes and reconstruct high-quality waveform even if the speaker is unseen in training. Compared to traditional audio codec approaches, the neural-based codec is significantly better at low bitrates, and we believe the quantized tokens contain sufficient information about the speaker and recording conditions. Compared to other quantization methods, the audio codec shows the following advantages: 1) It contains abundant  speaker information and  acoustic information, which could maintain speaker identity in reconstruction compared to HuBERT codes \citep{hsu2021hubert}. 2)  There is an off-the-shelf codec decoder to convert  discrete tokens into a waveform, without the additional efforts on vocoder training like VQ-based methods that operated on spectrum \citep{DBLP:conf/interspeech/DuGC022}. 3) It could reduce the length of time steps for efficiency to address the problem in $\mu$-law transformation \citep{DBLP:conf/ssw/OordDZSVGKSK16}. 

We adopt a pre-trained neural audio codec model, EnCodec \citep{defossez2022high}, as our tokenizer. EnCodec is a convolutional encoder-decoder model, whose input and output are both 24 kHz audio across variable bitrates. 
The encoder produces embeddings at 75 Hz for input waveforms at 24 kHz, which is a 320-fold reduction in the sampling rate. Each embedding is modeled by a residual vector quantization (RVQ), in which we choose eight hierarchy quantizers with 1024 entries each as shown in Figure \ref{codec}. This configuration corresponds to EnCodec at 6K bitrates for 24 kHz audio reconstruction. In this setting, given a 10-second waveform, the discrete representation is a matrix with $ 750 \times 8$ entries, where $750=\frac{24,000 \times 10}{320}$ is the downsampled time step and 8 is the number of quantizers. 
It is fine to choose other bitrate settings. A larger bitrate corresponds to more quantizers and better reconstruction quality. For example, if we choose EnCodecc at 12K bitrates, there are 16 quantizers are needed and the 10-second waveform corresponds to a matrix with $ 750 \times 16$ entries.  With the discrete codes from all quantizers, the convolutional decoder of EnCodec generates real-valued embeddings and reconstructs the waveform at 24 kHz. 

\section{\our{}}

\subsection{Problem Formulation: Regarding TTS as Conditional Codec Language Modeling}
Given a dataset $\mathcal{D} = \{\mathbf{x}_i , \mathbf{y}_i \}$, where $\mathbf{y} $ is an audio sample and $\mathbf{x} = \{ x_0, x_1, \ldots, x_L \}$ is its corresponding phoneme transcription, we use a pre-trained neural codec model to encode each audio sample into discrete acoustic codes, denoted as $ \text{Encodec}(\mathbf{y}) = \mathbf{C}^{T \times 8}$, where $\mathbf{C}$ represents the two-dimensional acoustic code matrix, and $T$ is the downsampled utterance length. The row vector of each acoustic code matrix $\mathbf{c}_{t,:}$ represents the eight codes for frame $t$ and the column vector of each acoustic code matrix $\mathbf{c}_{:, j}$ represents the code sequence from the $j$-th codebook, where $j \in \{1, \ldots , 8\}$. After quantization, the neural codec decoder is able to reconstruct the waveform, denoted as  $ \text{Decodec}(\mathbf{C})\approx \hat{\mathbf{y}}$. 

Zero-shot TTS requires the model to synthesize high-quality speech for unseen speakers. 
In this work, we regard zero-shot TTS as a conditional codec language modeling task. We train a neural language model to generate an acoustic code matrix $\mathbf{C}$ conditioned on a phoneme sequence $\mathbf{x}$ and  an acoustic prompt matrix $\mathbf{\tilde{C}}^{T'\times 8}$ with the optimization objective of $\max p (\mathbf{C} |\mathbf{x}, \mathbf{\tilde{C}})$. Here, $\mathbf{\tilde{C}}$ is obtained by the same neural codec with an enrolled recording as the input. We expect the  neural language model learns to extract the content and speaker information from the phoneme sequence and the acoustic prompt, respectively.
During inference, given a phoneme sequence and a 3-second enrolled recording of the unseen speaker, the acoustic code matrix with corresponding content and speaker's voice is firstly estimated by the trained language model. Then the neural codec decoder synthesizes the high-quality speech.


%
\begin{figure*}[t]
	\centering
	\includegraphics[width=\textwidth]{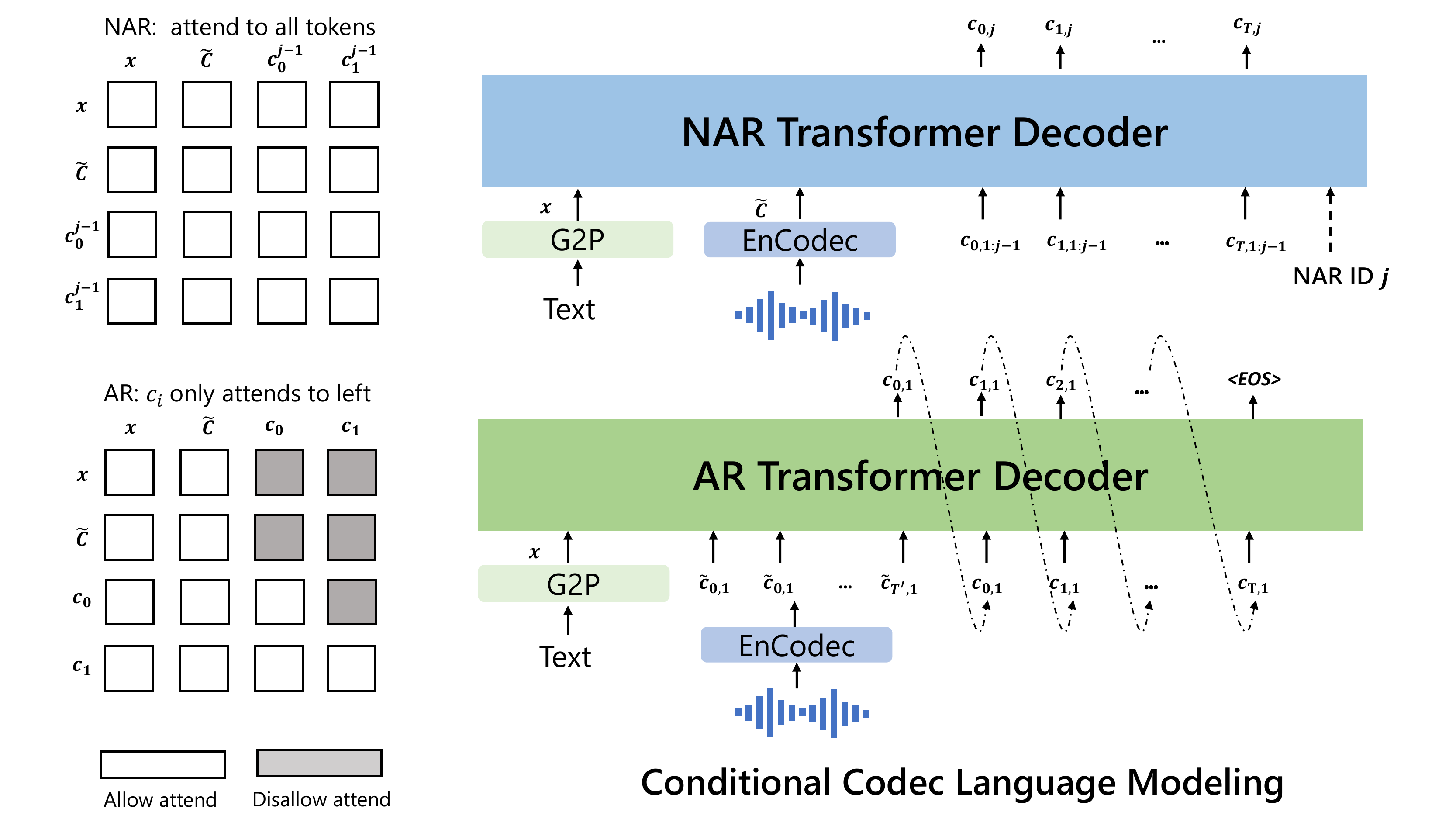}
	\caption{The structure of the conditional codec language modeling, which is built in a hierarchical manner. In practice, the NAR decoder will be called seven times to generate codes in seven quantizers. }\label{framework}
\end{figure*}%
\subsection{Training: Conditional Codec Language Modeling}

The neural speech codec model allows us to operate on discrete audio representations.  Due to residual quantization in the neural codec model, the tokens have a hierarchical structure: tokens from previous quantizers recover acoustic properties like speaker identity, while the consecutive quantizers learn fine acoustic details. Each quantizer is trained to model the residual from the previous quantizers. Motivated by this, we design two conditional language models in a hierarchical manner. 

For the discrete tokens from the first quantizer $\mathbf{c_{:, 1}}$, we train an autoregressive (AR) decoder-only language model. It is conditioned on the phoneme sequence $\mathbf{x}$ and the acoustic prompt $\mathbf{\tilde{C}}_{:,1}$, formulated as 
\begin{equation}
\label{AR}
    p (\mathbf{c}_{:,1} | \mathbf{x}, \mathbf{\tilde{C}}_{:,1}; \theta_{AR}) = \prod_{t=0}^{T} p(\mathbf{c}_{t, 1}|\mathbf{ c}_{<t, 1}, \mathbf{\tilde{c}}_{:,1}, \mathbf{x}; \theta_{AR})
\end{equation} 
Since \our{} is a decoder-only LM, the concatenation of $\mathbf{\tilde{c}_{:, 1}} $ and $\mathbf{c}_{:,1} $ is a whole sequence, and we do not distinguish them or insert a specific token in training. Only $\mathbf{c}_{:,1} $ is predicted while the prefix   $\mathbf{\tilde{c}}_{:,1} $ is given during inference.

For the discrete tokens from the second to the last quantizers, $\mathbf{c}_{:, j\in[2,8]}$, we train a non-autoregressive (NAR) language model. Since the tokens can not access each other in a NAR manner, to constrain the speaker identity, the acoustic prompt matrix $\tilde{C}$ is used as an acoustic prompt. Thus, the model is conditioned on the phoneme sequence $\mathbf{x}$, the acoustic prompt $\mathbf{\tilde{C}}$ and the predicted acoustic tokens belong to the previous codebooks $\mathbf{C}_{:, <j}$:
\begin{equation}
    p(\mathbf{C}_{:, 2:8}|\mathbf{x}, \mathbf{\tilde{C}}; \theta_{NAR}) =  \prod_{j=2}^{8} p(\mathbf{c}_{:,j} | \mathbf{C}_{:, <j}, \mathbf{x}, \mathbf{\tilde{C}}; \theta_{NAR})
\end{equation}

The combination of the AR model and the NAR model provides a good trade-off between speech quality and inference speed. On the one hand, the rate of the generated speech should be consistent with the enrolled recording, and it is hard to train a length predictor for different speakers since their speaking speed may be very diverse. In this case, the AR model is a more natural choice with its flexibility for acoustic sequence length prediction. On the other hand, for the consecutive stages, as the number of output slots follows the sequence length of the first stage, NAR can reduce the time complexity from $\mathcal{O}(T)$ to $\mathcal{O}(1)$. Overall, the prediction of $\mathbf{C}$ can be modeled as: 
\begin{equation}
    p(\mathbf{C}|\mathbf{x}, \mathbf{\tilde{C}}; \theta) = 
    p(\mathbf{c}_{:, 1}| \mathbf{\tilde{C}}_{:,1}, \mathbf{X};\theta_{AR}) 
    \prod_{j=2}^{8}p(\mathbf{c_{:, j}}|\mathbf{c}_{:, <j},\mathbf{x}, \mathbf{\tilde{C}}; \theta_{NAR})
\end{equation}

\subsubsection{Autoregressive Codec Language Modeling}
The autoregressive language model generates the tokens from the first quantizer. It comprises a phoneme embedding $W_x$, an acoustic embedding $W_a$, a transformer decoder, and a prediction layer. In order to generate speech with specific content, we use the phoneme sequence as the phoneme prompt of the language model. Thus, the model input is the concatenation of $\mathbf{x}$ and $\mathbf{c}_{:,1}$, and two special <EOS> tokens are appended after each of them. 
We compute sinuous position embedding separately for prompt and input tokens. For the causal transformer model, each token $c_{t,1}$ can attend to $(\mathbf{x}, \mathbf{c}_{\leq t,1})$ as illustrated in the left part of Figure \ref{framework}. The model is optimized to  maximize the probability of the next token in the first codebook. We share  the parameters of the output projection layer with the parameters of the acoustic embedding $W_a$.
 
In the AR model, we do not explicitly extract an audio clip as the prompt in training. The training process is pure casual language model training. In this way, any prefix sequence $\mathbf{c}_{<t,1}$ is treated as a prompt for the latter part of the sequence $\mathbf{c}_{\ge t, 1}$.  During inference, given an enrolled recording, we should concatenate the phoneme sequence of the enrolled recording and the phoneme sequence for synthesis together. Meanwhile, the acoustic token sequence of the enrolled recording is used as the prefix in AR decoding, as formulated in equation \ref{AR}. We will study the superiority of this setting in the experiment. 



\subsubsection{Non-Autoregressive Codec Language Modeling}
When we obtain the first quantizer codes by the AR model, we employ 
 a non-autoregressive (NAR) model   to generate codes of the other seven quantizers.  
The NAR model has a similar architecture to the AR model, except that it contains eight separate acoustic embedding layers. In each training step, we randomly sample a training stage $i \in [2, 8]$. The model is trained to maximize the acoustic tokens from the $i$-th quantizer codebook. The acoustic tokens from stage 1 to stage $i-1$ are embedded and summed up as model input:
\begin{small}
\begin{align}
    e_{c_{t,j}} &= W_a^j \odot {c_{t,j}} \\
    \mathbf{e_{c_t}} &= \sum_{j=1}^{i-1} e_{c_t,j}
\end{align}
\end{small}where $\odot$ indicates index selection.  The phoneme sequence is also regarded as the prompt of the language model. Besides, to clone the unique voice of the given speaker, we also use the acoustic tokens from the enrolled speech as the acoustic prompt. Specifically, we first tokenize the enrolled speech with the neural codec model as $\mathbf{\tilde{C}}^{T\times8}$. The embedded representations from all of the eight codebooks are summed up as the acoustic prompt $\mathbf{e_{\tilde{c}_t}} = \sum_{j=1}^{8} e_{\tilde{c}_{t,j}}$.
To predict the acoustic tokens from the $i$-th codebook, the transformer input is the concatenation of $(\mathbf{e_x}, \mathbf{e_{\tilde{c}}}, \mathbf{e_{c_{:,<i}}})$. The positional embeddings are also computed separately for prompts and the acoustic sequence. 
 The current stage $i$ is injected into the network with Adaptive Layer Normalization \citep{DBLP:conf/nips/Xu0ZZL19} operator, i.e., $\text{AdaLN}(h, i) = a_i\text{LayerNorm}(h)+b_i$, where $h$ is the intermediate activations, $a_i$ and $b_i$ are obtained from a linear projection of the stage embedding. Unlike AR, the NAR model allows each token to attend to all the input tokens in the self-attention layer.
 We also share the parameters of the acoustic embedding layer and the output prediction layer, which means the weights of the $j$-th prediction layer are the same as the $(j+1)$-th acoustic embedding layer.


\subsection{Inference: In-Context Learning via Prompting}

 In-context learning is a surprising ability of the text-based language model, which is able to predict labels for unseen inputs without additional parameter updates. For TTS, if the model can synthesize high-quality speech for unseen speakers without  fine-tuning, the model is believed to have in-context learning capability. However, the in-context learning capability of existing TTS systems is not strong, because they  either require additional fine-tuning or degrade dramatically for unseen speakers.  

 For language models, prompting is necessary to enable in-context learning in the zero-shot scenario. We design prompts and inference as follows. 
 We first convert the text into  a phoneme sequence  and encode the enrolled recording into an acoustic matrix, forming the phoneme prompt and acoustic prompt. Both prompts are used in the AR and NAR models.  For the AR model, we use sampling-based decoding conditioned on the prompts  since we observe that beam search may lead the LM into an infinity loop. Furthermore, the sampling-based method could significantly increase the diversity of the output.  For the NAR model, we use greedy decoding to choose the token with the highest probability. Finally, we use the neural codec decoder to generate the waveform conditioned on the eight code sequences. The acoustic prompt may or may not  semantically relate to the speech to be synthesized, resulting in two cases:

\noindent \textbf{\our{}}: Our main interest is to generate given content for unseen speakers. The model is given a text sentence, a segment of enrolled speech, and its corresponding transcription. 
We prepend the transcription phoneme of the enrolled speech to the phoneme sequence of the given sentence as the phoneme prompt, and use the first layer acoustic token of the enrolled speech $\tilde{c}_{:, 1}$ as an acoustic prefix. With the phoneme prompt and the acoustic prefix, \our{} generates the acoustic tokens for the given text cloning the voice of this speaker. 

\noindent \textbf{\our{}-continual}:  In this setting, we use the whole transcription and the first 3 seconds of the utterance as the phoneme and acoustic prompts respectively, and ask the model to generate the continuations. The inference process is the same as setting \our{}, except that the enrolled speech and the generated speech are semantically continuous.

\section{Experiment}
\subsection{Experiment Setup}

\textbf{Dataset}: We use LibriLight \citep{kahn2020libri} as the training data which contains 60K hours of unlabelled speech from audiobooks in English. The number of distinct speakers is around 7000 in  LibriLight.  We train a hybrid DNN-HMM ASR model on 960 hours labeled LibriSpeech following Kaldi recipe~\citep{povey2011kaldi}. 
Once the hybrid model is trained, unlabeled speech data is decoded and transduced to the best phoneme-level alignment paths where the frameshift is 30ms. The EnCodec model \citep{defossez2022high} is used to generate the acoustic code matrix for the 60K hours of data. 

\textbf{Model}: Both the AR model and the NAR model have the same transformer architecture with 12 layers, 16 attention heads, an embedding dimension of 1024, a feed-forward layer dimension of 4096, and a dropout of 0.1. The average length of the waveform in LibriLight is 60 seconds. During training, we randomly crop the waveform to a random length between 10 seconds and 20 seconds. Its corresponding phoneme alignments are used as the phoneme prompt. We remove the consecutive repetitions in the force-aligned phoneme sequence. 
For the NAR acoustic prompt tokens, we select a random segment waveform of 3 seconds from the same utterance.

 The models are trained using 16 NVIDIA TESLA V100 32GB GPUs with a batch size of 6k acoustic tokens per GPU for 800k steps. We optimize the models with the AdamW optimizer, warm up the learning rate for the first 32k updates to a peak of $5 \times 10^{-4}$, and then linear decay it. 

 \textbf{Baseline}:  We choose the SOTA zero-shot TTS model YourTTS \citep{casanova2022yourtts} as the baseline, which is trained on a combined dataset of VCTK \citep{veaux2016superseded}, LibriTTS \citep{DBLP:conf/interspeech/ZenDCZWJCW19}, and TTS-Portuguese \citep{DBLP:journals/lre/CasanovaJSOTPA22}. We use their released checkpoint\footnote{\url{https://github.com/Edresson/YourTTS}}. 

 \textbf{Automatic metrics}: We employ the SOTA speaker verification model, WavLM-TDNN \citep{chen2022wavlm}, to evaluate the speaker similarity between prompt (the decompressed enrolled speech) and synthesized speech. WavLM-TDNN achieved the top rank at the VoxSRC Challenge 2021 and 2022 leaderboards. It reached an average Equal Error
Rate (EER) of 0.383, 0.480, and 0.986 on Vox1-O, Vox1-E, and Vox1-H respectively. The similarity score predicted by WavLM-TDNN is in the range of $[-1,1]$, where a larger value indicates a higher similarity of input samples. 

We also evaluate the synthesis robustness of our model. Neural TTS systems suffer from the robustness issue, which sometimes has deletion, insertion, and replacement errors due to wrong attention alignments.  We perform ASR on the generated audio and calculate the word error rate (WER) with respect to the original transcriptions. In this experiment, we employ the HuBERT-Large \citep{hsu2021hubert} model fine-tuned on LibriSpeech 960h as the ASR model, which is a CTC-based model without language model fusion. 

 \textbf{Human evaluation}: We calculate the comparative mean option score (CMOS) and similarity mean option score (SMOS)  by  crowdsourcing, where 12 and 6 native speakers are invited as CMOS and SMOS contributors. The scale of SMOS is from 1 to 5 with 0.5-point increments.  CMOS ranges from -3 (the new system is
much worse than baseline) to 3 (the new system is much better than
baseline) with intervals of 1.
CMOS is an indicator of speech naturalness, and SMOS measures whether the speech is similar to the original speaker's voice.

\subsection{LibriSpeech Evaluation}
We first use LibriSpeech \citep{panayotov2015librispeech} for zero-shot TTS evaluation, since there is no speaker overlap between LibriLight training data and LibriSpeech test-clean data.
Following \cite{DBLP:journals/corr/abs-2209-03143},  we use the samples from LibriSpeech test-clean with lengths between 4 and 10 seconds, resulting in a 2.2 hours subset. 
For each sample synthesis, \our{} randomly choose another utterance of the same speaker and crop a 3-seconds speech segment as the enrolled speech. Each experiment runs three times and the average score is reported. \our{}-continual uses the first 3 seconds of the ground-truth speech as enrolled speech. 
\begin{table}[h]
\begin{center}
\caption{Evaluation results on audio generation. YourTTS and \our{} are text-to-speech models using phonemes as inputs, while GSLM and AudioLM are speech-to-speech models using latent code as inputs.  The WER result of AudioLM is obtained by a Conformer Transducer model \citep{DBLP:journals/corr/abs-2209-03143}. Since AudioLM* is not open-source, we cannot evaluate its speaker score with our tool.  }
\begin{tabular}{lcc}
\toprule
model         & WER & SPK \\ \hline
GroundTruth   &  2.2  & 0.754  \\  \hline
 \multicolumn{3}{l}{\textbf{Speech-to-Speech Systems}} \\ 
GSLM            &   12.4    &   0.126 \\ 
AudioLM*         &  6.0  & -   \\ \hline
 \multicolumn{3}{l}{\textbf{TTS Systems}}  \\ 
YourTTS           &   7.7    &  0.337  \\ 
\our{}            &  5.9   &  \bf 0.580  \\ 
\our{}-continual  &  \bf 3.8  &   0.508 \\ \hline
\end{tabular}
\label{exp:librispeech}
\end{center}
\end{table}

Table \ref{exp:librispeech} shows the objective evaluation results. We first compute the WER score and the speaker similarity score of the ground truth speech as the upper bound. To compare the speaker similarity, we use speech pairs from the same speaker in the test set. Compared with the YourTTS baseline, our model is significantly better in both robustness and speaker similarity, showing that our generated speech is highly faithful to the given text and the given enrolled speech. Furthermore, the word error rate can be further reduced in \our{}-continual setting, because the acoustic tokens for the first 3 seconds are extracted from the ground truth.  We also compare the robustness with other speech-to-speech LM-based generation models, GSLM and AudioLM, which use audio latent codes as input. GSLM uses HuBERT code as input and reconstructs the waveform with the Tacotron2 \citep{DBLP:conf/icassp/ShenPWSJYCZWRSA18} model and the WaveGlow \citep{prenger2019waveglow} vocoder. We run their open-sourced code using the released model and evaluate the results. Since the HuBERT codes discard the speaker identity, it achieves a poor speaker score.  For the AudioLM, we list their WER score reported in their paper, which is obtained by a Conformer Transducer model.  The experiment results show that \our{} is better than other speech-to-speech LM-based generative systems in  terms of robustness. One major reason is \our{} trained with pseudo-phoneme instead of HuBERT/w2v-BERT codes, which enjoys better alignment quality with the input text.

We randomly sample one utterance for each speaker in LibriSpeech test-clean for the human evaluation, resulting in 40 test cases. Table \ref{exp:libri_spk_human} shows the human evaluation results. \our{} is very closed to ground truth in terms of SMOS, indicating the synthesized speech is similar to the given unseen speaker in testing. It significantly outperforms the baseline with +0.93 SMOS, demonstrating the effectiveness of \our{} in zero-shot scenarios. 
Regarding naturalness, \our{} beats the baseline with +0.12 CMOS, indicating the proposed method could synthesize more natural and realistic speech against baselines. 

\begin{table}[h]
\begin{center}
\caption{Human evaluation with 40 speakers on LibriSpeech test-clean with 3-second enrolled recording for each. \label{exp:libri_spk_human} }
\begin{tabular}{lcc}
\toprule
& SMOS & CMOS (v.s. \our{})   \\
\hline
YourTTS  & 3.45$_{\pm 0.09}$ & -0.12 \\ 
\our{} &  4.38$_{\pm 0.10}$ & 0.00\\ 
GroundTruth &   4.5$_{\pm 0.10}$ & +0.17\\ 
 \bottomrule
\end{tabular}

\end{center}
\end{table}

\textbf{Ablation study:}
In this section, we perform detailed ablation experiments. We first study the NAR model. We train three NAR models with different numbers of prompts. The setting \textbf{NAR-no prompt} is trained without any prompts. The setting \textbf{NAR-phn prompt} is trained with only phoneme sequence as prompt and the setting \textbf{NAR-2 prompts} uses both phoneme prompt and acoustic token prompt as conditions. In evaluation, we use the ground-truth first-level acoustic tokens as the model input and compute the WER and speaker similarity scores. The results are listed in Table \ref{exp:NAR}. 
\begin{table}[h]
\begin{center}
\caption{Ablation study of the NAR model. The inputs of the NAR models are the ground-truth for the ablation study.   }
\begin{tabular}{cccc}
\toprule
& \bf NAR-no prompt & \bf NAR-phn prompt & \bf NAR-2 prompts \\  \hline
WER & 19.6 & 3.0  & \textbf{2.8} \\ \hline
SPK & 0.518 & 0.541  & \textbf{0.732} \\ \hline
\end{tabular}
\label{exp:NAR}
\end{center}
\end{table}
Results show that the model without any prompts performs poorly on both ASR and speaker similarity evaluations, even though the acoustic input token is ground truth. When adding the phoneme  prompt, the WER is reduced by a large margin from 19.6 to 3.0. It shows the phoneme prompt mainly contributes to the content of the generation. In the \textbf{NAR-2 prompts}, the  model can learn speaker information from the acoustic token prompt and thus improve the speaker evaluation quality.

We further conduct the ablation experiments on the AR model. In these experiments, we always use the \textbf{NAR-2 prompts} setting as the NAR model.   In Table \ref{exp:AR}, we can see that when we remove the acoustic prompt (w/o acoustic prompt), it can only obtain a speaker similarity score of 0.236, showing the prompt is extremely crucial for speaker identity. Even if the NAR model could see the prompt, the prompt for the AR model also contributes a lot to speaker similarity.  

\begin{table}[h]
\begin{center}
\caption{Ablation study of the AR model.  }
\begin{tabular}{ccc}
\toprule
   & WER & SPK  \\  \hline
    \our{}  &  \textbf{5.9}  &  \textbf{0.585} \\ 
   w/o acoustic prompt   & \textbf{5.9} & 0.236 \\ \hline
\end{tabular}
\label{exp:AR}
\end{center}
\end{table}



\subsection{VCTK Evaluation}
We evaluate our model on VCTK consisting of 108 speakers, where none of the speakers are observed during training. Since YourTTS has seen 97 speakers in VCTK as training, we evaluate YourTTS performance on the full 107 speakers and 11 unseen speakers, respectively. 
For each speaker, we randomly selected three utterances of 3s/5s/10s as the prompts and the text of another utterance as the text prompt.

\begin{table}[h]
\begin{center}
\caption{Automatic evaluation of speaker similarity with 108 speakers on VCTK. *YourTTS has observed 97 speakers during training, while \our{} observed none of them. \label{exp:vctk_spk}}
\begin{tabular}{lccc}
\toprule
& 3s prompt & 5s prompt & 10s prompt  \\
\hline
\multicolumn{4}{c}{108 full speakers} \\
\hline
YourTTS* & 0.357 &   0.377 &  0.394 \\ 
\our{} & 0.382 &   0.423  & \textbf{0.484}  \\ 
GroundTruth & 0.546 &  0.591 &  0.620   \\ 
\hline
\multicolumn{4}{c}{11 unseen speakers} \\
\hline
YourTTS  & 0.331 &  0.337 &   0.344 \\ 
\our{} & 0.389 &  0.380 &   \textbf{0.414}   \\ 
GroundTruth & 0.528 &  0.556 &   0.586  \\ 
         \bottomrule
\end{tabular}

\end{center}
\end{table}
We first evaluate two models with the speaker verification metric as described before. 
From Table \ref{exp:vctk_spk}, we can see that \our{} outperforms the baseline even if the baseline has seen 97 speakers in training, indicating our model is able to synthesize speech with higher speaker similarity. When we compare with the baseline in a fair setting (11 speakers), the performance gap becomes larger, especially when only 3s prompts are available. By comparing different lengths of the prompt, we can see our model is able to generate more similar speech when the prompt becomes longer, which is consistent with our intuition.  


\begin{table}[h]
\begin{center}
\caption{Human evaluation with 60 speakers on VCTK with 3-second enrolled recording for each.  }
\begin{tabular}{lcc}
\toprule
& SMOS & CMOS (v.s. \our{})   \\
\hline
YourTTS* & 3.70$_{\pm 0.09}$ &   -0.23 \\ 
\our{} & 3.81$_{\pm 0.09}$ &   0.00    \\ 
GroundTruth & 4.29$_{\pm 0.09}$ &  -0.04    \\ 
 \bottomrule
\end{tabular}
\label{exp:vctk_spk_human}
\end{center}
\end{table}

We sample 60 speakers for human evaluation, one utterance for each, where 11 are unseen speakers, and 49 speakers have been seen for YourTTS. \our{} do not see any of the 60 speakers. During model synthesis, each speaker has a 3-second enrolled recording. Table \ref{exp:vctk_spk_human} shows a comparison of our method against baseline and ground truth. The comparison of SMOS shows that \our{} has better speaker similarity than the baseline, even if the baseline has seen some of the speakers in training. 
The side-by-side CMOS evaluation shows that \our{} is +0.23 over YourTTS, indicating a significantly better performance on speaking of naturalness. Furthermore, \our{} achieves +0.04 CMOS over ground-truth, demonstrating no statistically significant difference from human recordings on this dataset. 
 Compared to the evaluation results on LibriSpeech, \our{} shows a better CMOS score in the comparison with ground truth, which is mainly because the average sentence length is shorter  and some of the ground truth utterances also have noisy environments in VCTK.  In terms of speaker similarity, VCTK is more challenging as it contains speakers with various accents while the training data and LibriSpeech test data do not contain various accent speakers. 
 \begin{figure}[!h]
	\centering
     \subfigure[A LibriSpeech sample: After early nightfall, the yellow lamp would light up here and there the squalid quarter of the brothels.
]{\includegraphics[width=0.85 \textwidth]{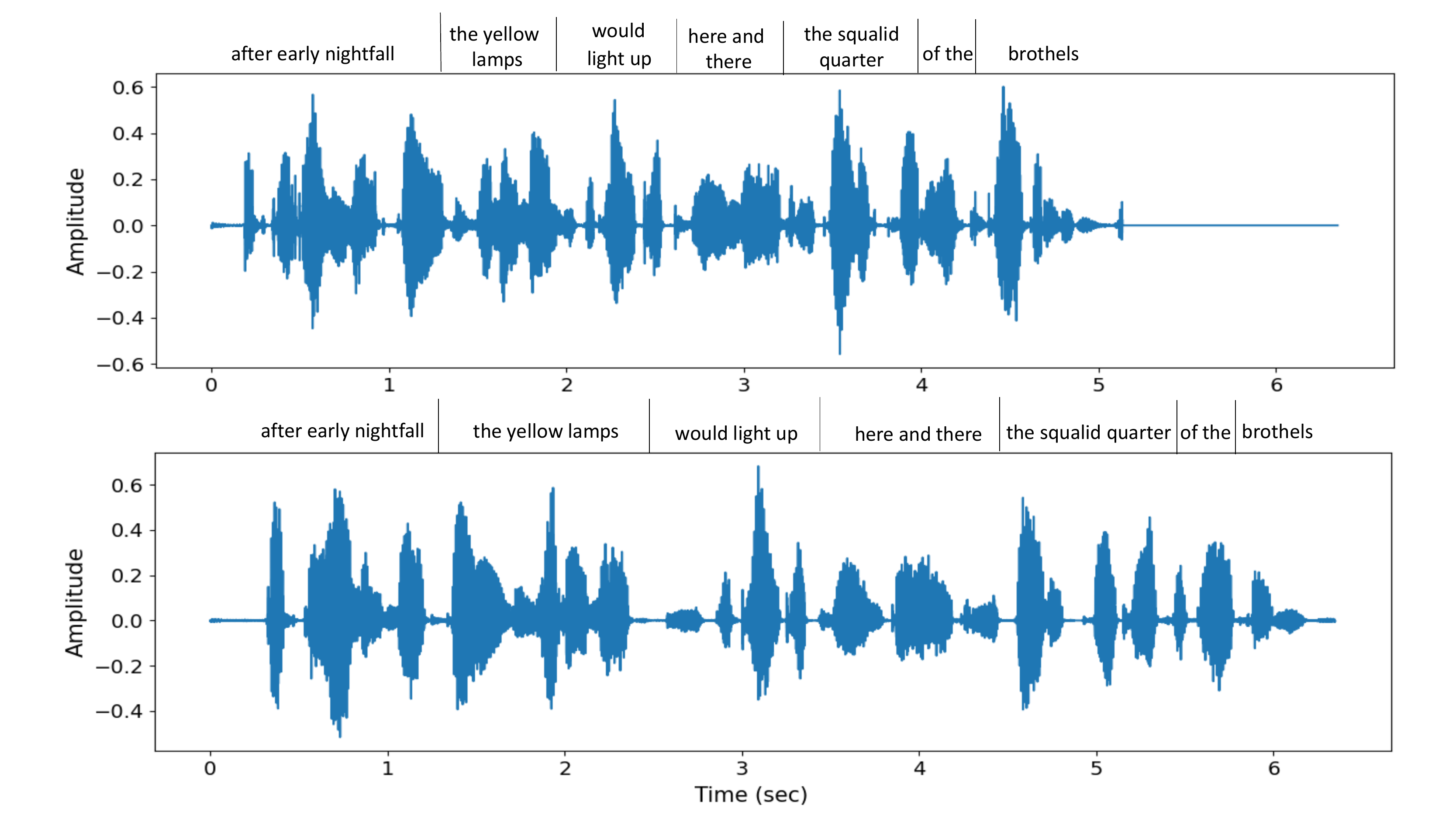}
     \label{fig:visual1}} 
	\subfigure[A VCTK sample: I must do something about it.]{\includegraphics[width=0.88 \textwidth]
     {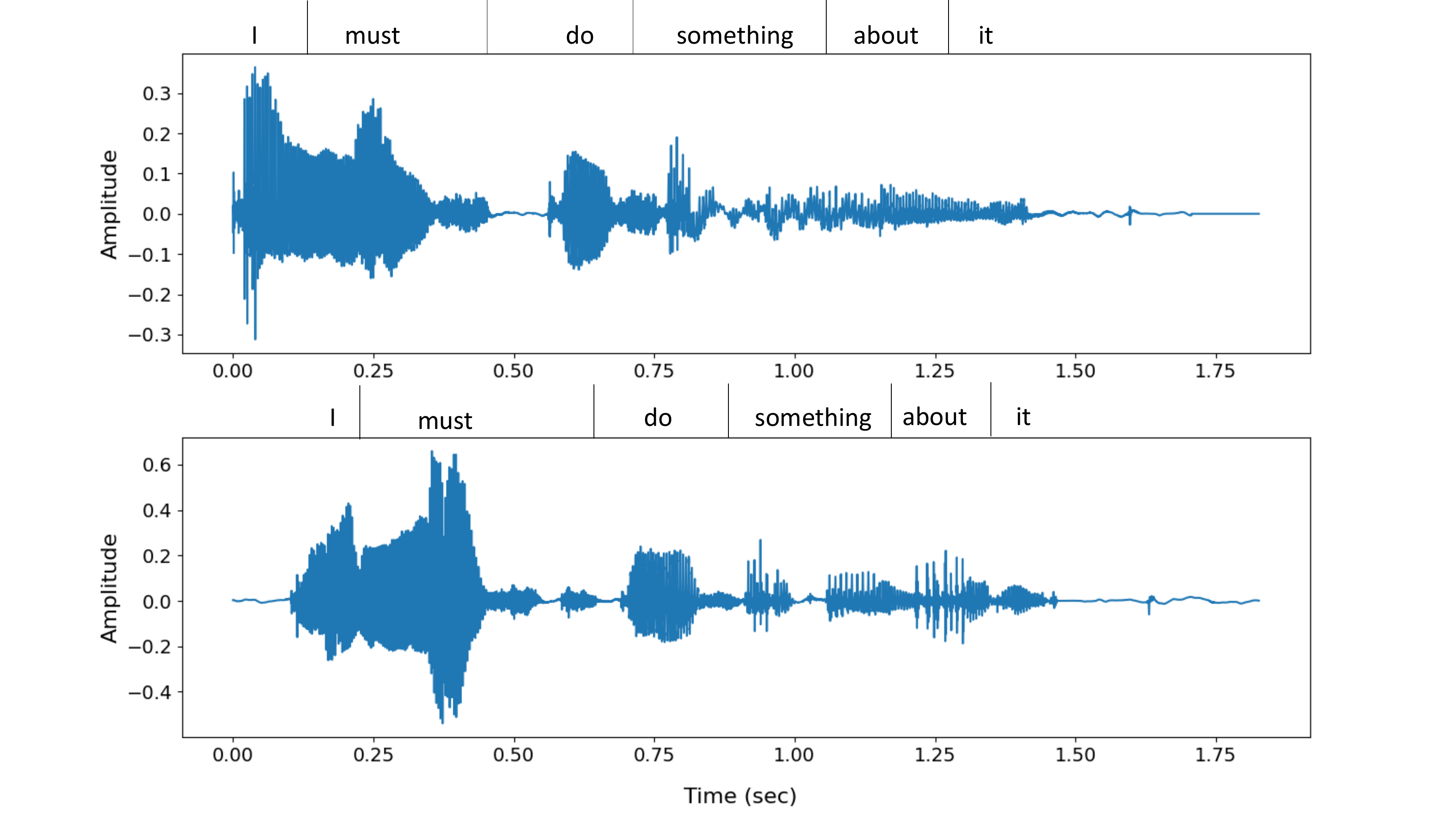}\label{fig:visual2}} 

	\caption{Diversity analysis of \our{}. Each utterance is synthesized two times with different random seeds. We can observe substantial diversity of the two outputs regarding the same input.  }
    \label{fig:visual}
    \centering
\end{figure}

\subsection{Qualitative Analysis}
\textbf{Diversity:} Previous TTS systems have a strong one-one mapping between input text and output waveform, because mel spectrum generation is based on reconstruction for each step without randomness. Since \our{} uses the sampling-based method to generate discrete tokens, its output is diverse for the same input text due to the randomness in inference. Given a sentence and an enrolled recording, we run the inference process twice and visualize its waveform in Figure   \ref{fig:visual}. In Figure \ref{fig:visual1}, we observe the two samples have different lengths and phrase durations, where the first has a faster speech rate. In Figure \ref{fig:visual2}, we observe that the accents of the two samples are different. The second output emphasizes the word ``must" with a larger amplitude whereas the first output does not. We leave more samples on our demo page. 

The diversity is important for some downstream scenarios. For example,  speech recognition always benefits from diverse inputs with different speakers and acoustic environments, which cannot be met by the previous TTS system. Considering the diversity feature of \our{}, it is an ideal candidate to generate pseudo-data for speech recognition. 

\textbf{Acoustic environment maintenance:} Another interesting finding is the acoustic environment consistency between the acoustic prompt and the generation. When the acoustic prompt has reverberation, \our{} could synthesize speech with reverberation as well, whereas the baseline outputs clean speech. Our explanation is that \our{} is trained on a large-scale dataset consisting of more acoustic conditions than the data used by the baseline, so \our{} could learn the acoustic consistency instead of a clean environment only during training.  We show consistency on our demo page. 

\textbf{Speaker's emotion maintenance:} Emotional TTS is a classic subtopic of speech synthesis, which synthesizes speech with a required emotion. Traditional methods \citep{lei2021fine} always train a model on a supervised emotional TTS dataset, where the speech corresponds to a transcription and an emotion label. We find that \our{} can preserve the emotion in the prompt at a zero-shot setting. We select acoustic prompts from EmoV-DB \citep{adigwe2018emotional}, a dataset containing speech with five emotions, \our{} is able to keep the same emotion of the prompt in speech synthesis, even if  the model is not fine-tuned on an emotional TTS dataset. We put audio samples on our demo page.

\section{Conclusion, Limitations, and Future Work}
We introduced \our{}, a language model approach for TTS with audio codec codes
as intermediate representations.
We pre-train \our{} with 60K hours of speech data, and show the in-context learning capability in zero-shot scenarios. We achieve new state-of-the-art zero-shot TTS results on LibriSpeech and VCTK. Furthermore, \our{} could keep the acoustic environment and speaker's emotion in synthesis, and provide diverse outputs in different sampling-based decoding processes.  

Despite making significant progress, \our{} still suffers from several issues. 

\textbf{Synthesis robustness:} We observe that some words may be unclear, missed, or duplicated in speech synthesis. It is mainly because the phoneme-to-acoustic language part is an autoregressive model, in which disordered attention alignments exist and no constraints to solving the issue. The phenomenon is also observed in vanilla Transformer-based TTS, which was addressed by applying non-autoregressive models or modifying the attention mechanism in modeling. In the future, we would like to leverage these techniques to solve the issue. 

\textbf{Data coverage:} Even if we use 60K hours of data for  training, it still cannot cover everyone's voice, especially accent speakers. The worse result on VCTK than LibriSpeech also implies  insufficient coverage of accent speakers.  Moreover, the  diversity of speaking styles is not enough, as LibriLight is an audiobook dataset, in which most utterances are in reading style.  In the future, we will further scale up the training data to improve the model performance across prosody, speaking style, and speaker similarity perspectives. We believe the zero-shot TTS task could be almost solved through our approach with model and data scale-up. 

\textbf{Model Structure:} Now, we use two models to predict codes of different quantizers. A promising direction is to predict them with a large universal model. Another interesting direction  is using full NAR models to speed up model inference  in the framework. 


\textbf{Broader impacts:} Since \our{} could synthesize speech that maintains speaker identity, it may carry potential risks in misuse of the model, such as spoofing voice identification or impersonating a specific speaker. To mitigate such risks, it is possible to build a detection model to discriminate whether an audio clip was synthesized by \our{}. We will also put Microsoft AI Principles\footnote{\url{ https://www.microsoft.com/ai/responsible-ai}} into practice when further developing the models.

\bibliography{neurips_2022}
\bibliographystyle{plainnat}
\end{document}